\documentclass[letterpaper, 10 pt, conference]{ieeeconf}  

\IEEEoverridecommandlockouts                              

\usepackage{graphicx}
\graphicspath{ {./images/} }
\usepackage[strings]{underscore} 

\usepackage{amsmath} 
\usepackage{amssymb}  
\usepackage{float}
\usepackage{fancyvrb}
\usepackage[colorlinks,urlcolor=blue]{hyperref}

\title{\LARGE \bf
A Game Benchmark for Real-Time Human-Swarm Control
}

\author{
\thanks{Authors are with the Center for Robotics and Biosystems at Northwestern University. \textbf{Corresponding Author Email:} joelmeyer@u.northwestern.edu}
Joel Meyer, Allison Pinosky, Thomas Trzpit, Ed Colgate, Todd D. Murphey
}

\begin{document}

\maketitle
\thispagestyle{empty}
\pagestyle{empty}

\begin{abstract}
We present a game benchmark for testing human-swarm control algorithms and interfaces in a real-time, high-cadence scenario. Our benchmark consists of a swarm vs. swarm game in a virtual ROS environment in which the goal of the game is to ``capture'' all agents from the opposing swarm; the game's high-cadence is a result of the capture rules, which cause agent team sizes to fluctuate rapidly. These rules require players to consider both the number of agents currently at their disposal and the behavior of their opponent's swarm when they plan actions. We demonstrate our game benchmark with a default human-swarm control system that enables a player to interact with their swarm through a high-level touchscreen interface. The touchscreen interface transforms player gestures into swarm control commands via a low-level decentralized ergodic control framework. We compare our default human-swarm control system to a flocking-based control system, and discuss traits that are crucial for swarm control algorithms and interfaces operating in real-time, high-cadence scenarios like our game benchmark. Our game benchmark code is available on Github; more information can be found at \url{https://sites.google.com/view/swarm-game-benchmark}.

\end{abstract}

\section{Introduction}

An appealing aspect of robot swarms is their potential to be deployed into areas that are dangerous for humans to enter. Many dangers---like fire and unstable structures---cause physical harm and evolve rapidly. Operators controlling robot swarms deployed into environments with these dangers cannot assume the environment will remain constant; they need to be able to rapidly adapt to changing conditions, new information, and swarm agent dropout. 

Many dangers are challenging to simulate because each environment is unique, however, simulations can still play an important role in testing human-swarm control algorithms and interfaces for these environments. Thus, the vision of sending human-swarm teams into dangerous, rapidly evolving scenarios faces two challenges: 1) many existing algorithms and interfaces are ill suited to the challenges of highly-dynamic environments and 2) there are no benchmarks (real-world or simulation-based) for testing such proposed algorithms and interfaces. To tackle these challenges, we present a game benchmark inspired by the challenges of real-time human-swarm control in high-cadence environments.

Even in benign scenarios, controlling swarms is challenging. As swarm size grows, the cognitive load on the operator increases, and it becomes difficult for an operator to task individual agents \cite{durantin_using_2014}. Prior work has sought to reduce the operator's cognitive load by autonomously planning agent trajectories via flocking algorithms \cite{olfati-saber_flocking_2006}, potential fields \cite{howard_mobile_2002}, fixed formations \cite{michael_planning_2009}, linear temporal logic \cite{chen_distributed_2021}, collective motion \cite{szwaykowska_collective_2015} \cite{zhao_self-adaptive_2018}, and Voronoi partitions \cite{sampedro_flexible_2016}. Unfortunately, many of these algorithms are too rigid for highly dynamic and dangerous scenarios and limit the variety of commands an operator can use to achieve a task. 
\begin{figure}
	\centering
	\includegraphics[width=3.4in]{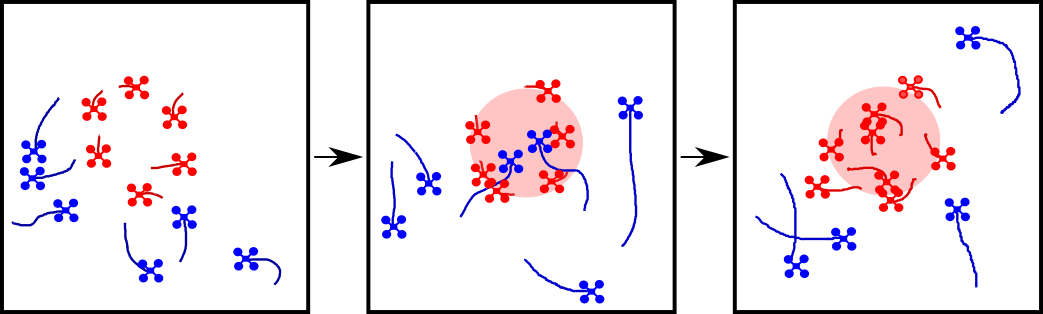}
	\caption{\textbf{Swarm vs. Swarm Game:} 
	Two players use interfaces to control opposing swarms in a shared, continuous space. The objective is to capture all agents on the opposing swarm by maneuvering one's own swarm to surround opposing agents. In this figure, as the blue swarm moves about, the red swarm forms a circular structure that surrounds the center of the game environment. Blue agents that pass through this surrounded area are captured by the red swarm according to the game rules. The game rules, which lend themselves to the emergence of vaguely Go-like structures during game play, are further explained in Section \ref{sec:game-rules}). The blue agents that are captured become part of the red swarm, and can now be controlled by the red player.}
	\label{fig:system-overview-cartoon}
	\vspace{-1em}
\end{figure}

Density-based swarm control algorithms, on the other hand, enable operators to specify flexible behavior to their swarm in dynamic and potentially dangerous scenarios. For example, authors in \cite{diaz-mercado_distributed_2015} developed a system that enabled human operators to control their swarm through density specifications via a touchscreen interface. Authors in \cite{prabhakar_ergodic_2020} created an end-to-end swarm control system that leveraged their swarm's heterogeneous capability, used autonomously detected information to keep the swarm safe, and enabled operators to specify multimodal commands to their swarm via touchscreen that were adaptable in real-time.
Furthermore, the touchscreen interfaces used in these works to send commands enabled both persistent swarm behavior and multiple variations of operator commands to achieve tasks through density specifications. Other work in swarm interfaces (e.g., touchscreen interfaces \cite{tsykunov_swarmtouch_2019}, brain-machine interfaces \cite{karavas_hybrid_2017}, swarm programming languages \cite{pinciroli_buzz_2016}, and haptic control devices \cite{lee_haptic_2011}), were more rigid in the types of behavior they could specify. Some of these interfaces were also not persistent and required the operator to constantly input commands to their system. In this work, we extend the system presented in \cite{prabhakar_ergodic_2020} to demonstrate our game benchmark due to the system's promising real-time performance.

\begin{figure}
    \vspace{1em}
	\centering
	\includegraphics[width=3.25in]{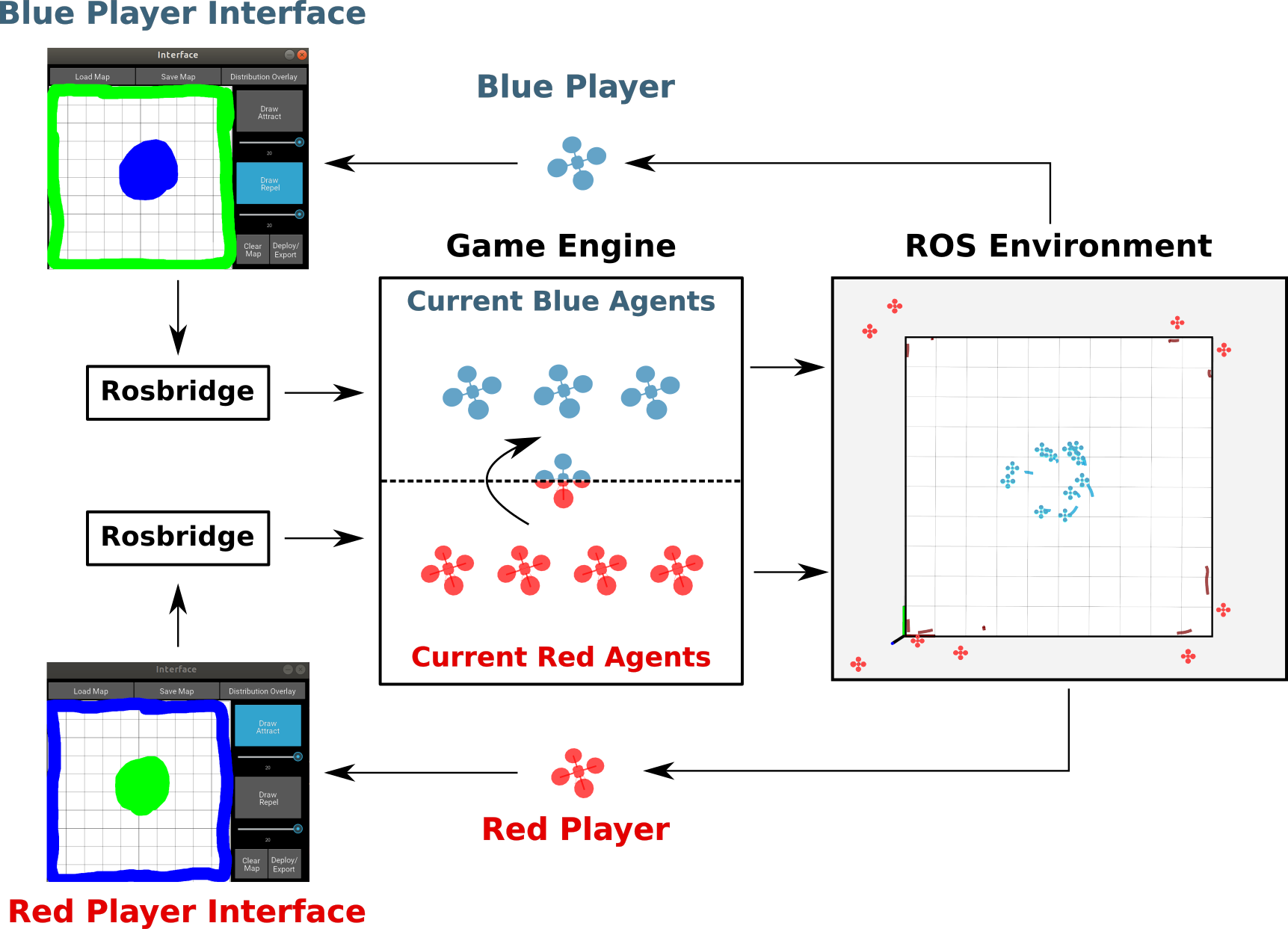}
	\caption{\textbf{Overview of the Game Benchmark:} Two players use separate interfaces (pictured here are the touchscreen interfaces for our human-swarm control system described in Section \ref{sec:system-overview}; researchers can choose to develop their own interfaces) to send commands via ROSbridge to their separate teams of agents. A ROS node tracks the state of the game (the positions of all agents) and determines whether an agent has been captured based on the rules of the game. Both players have their own displays (in RViz) showing the current state of the game. All code for the game benchmark will be made open source and can be executed on any laptop running Ubuntu 18.04 LTS with an Intel I5 or equivalent processor, and 4 GB of RAM. The code will contain a touchscreen interface for sending commands and an implementation of decentralized ergodic control for executing commands to enable researchers to demo the game benchmark ``out of the box''.}
	\label{fig:testbed-architecture}
	\vspace{-1em}
\end{figure}

Prior work involving swarm testbeds and real-world swarm demonstrations has enabled researchers to test formation control, motion planning, and collision avoidance \cite{preiss_crazyswarm_2017}, \cite{michael_grasp_2010}, \cite{pickem_robotarium_2017}, \cite{kate_simbeeotic_2012}, \cite{chung_live-fly_2016}. However, none of these testbeds or demonstrations were particularly dynamic nor did they assess high-cadence swarming. Work in game-theoretic scenarios involving pursuit-evasion and racing with multi-agent systems \cite{wang_game-theoretic_2021}, \cite{shah_multi-agent_2019} \cite{shah_grape_2019}, \cite{pierson_intercepting_2017}, \cite{pierson_controlling_2018} has created scenarios and algorithms that have approached what is needed to assess real-time swarm control in dynamic, high-cadence environments. The authors of \cite{day_responding_2018} conducted a field test with unmanned aerial vehicles involving teams of 10 vs. 10 agents in which they examined swarm combat tactics in a mock aerial dog fight. However, all of these works were limited to team sizes under 10 agents, and some did not involve human operators.

Given the lack of systems and testing scenarios for real-time high-cadence human-swarm control, we have created a game benchmark (shown in Figure \ref{fig:system-overview-cartoon}) that consists of a ROS-based virtual environment containing a swarm vs. swarm game. The swarm vs. swarm game targets scenarios that require evolving strategies---that is, rapid re-specification of strategies---in a rapidly changing environment (with respect to agent positions). We demonstrate our game benchmark with a human-swarm control system that uses a touchscreen interface to specify gesture-based objectives as distributions for robot swarms using real-time, decentralized ergodic control. 

Section~\ref{sec:benchmark} describes our dynamic swarm vs. swarm game benchmark. Section~\ref{sec:system-overview} describes the default system for real-time human-swarm control we use to demonstrate our game. Section~\ref{sec:swarm-tactics} describes an example tactic that can be deployed in the game with our default system, highlights considerations operators may need to make when planning tactics in the game, compares our default system for demonstrating the game to a flocking-based method, and discusses the shortcomings of other swarm control systems for planning tactics in high-cadence scenarios. In Section \ref{sec:discussion}, we discuss traits of real-time human-swarm control algorithms and interfaces that may be desirable for high-cadence scenarios like our game benchmark and conclude with future work. 

\section{The Game Benchmark}
\label{sec:benchmark}
In this section, we describe our benchmark, which is a dynamic game in which players control their swarms to capture \emph{all} of the agents on the opposing team. Both teams start with the same number of agents, which is specified by the players when the game is initialized. The game is played until all the available agents are captured by one team (i.e., if the game starts with 20 agents total, 10 on each team, the game does not end until one team has all 20 agents and the other team has 0). 

Players must strategically maneuver their swarm to surround agents on the opposing swarm while preventing their own agents from being captured---resulting in the emergence of vaguely Go-like structures (see Figure \ref{fig:system-overview-cartoon}). We note, however, that our game is played in continuous time and space---there is no notion of teams ``taking turns''. Also, all of the agents in the game can be at any location in the environment at any time. Figure \ref{fig:testbed-architecture} shows the architecture for our game benchmark. The game benchmark architecture includes the agents, the control algorithm and interface the players choose to deploy, the virtual ROS environment, and the game engine dictating the capture rules.

\begin{figure*}
    \vspace{1em}
	\centering
	\includegraphics[width=6.5in]{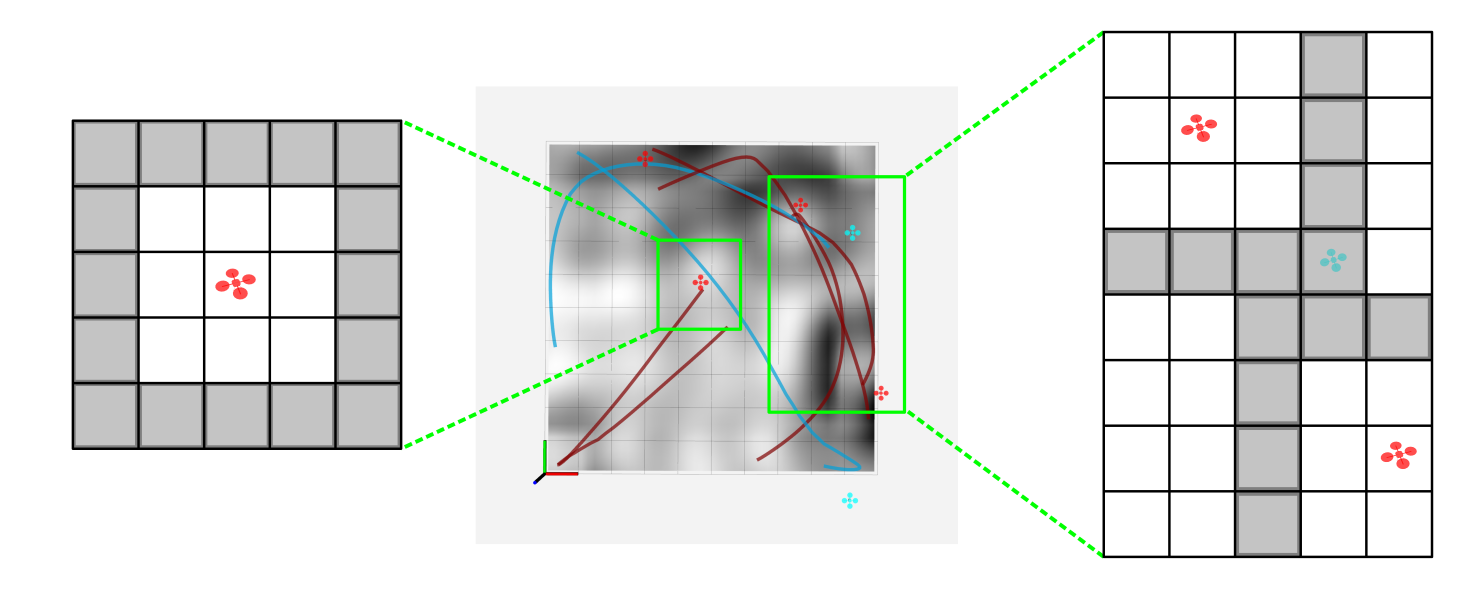}
	\caption{\textbf{Swarm-based Game Geometry:} Each agent contributes to the time-averaged trajectories of the team it is on through its individual trajectory history. At each instance in time, the area surrounding an agent's current position is weighted as a capture area for its team (as seen in the pop-out on the left). Several agents with overlapping capture areas are more likely than an individual agent to capture an agent on the opposing team. The combined capture areas of the two red agents (visualized in the pop-out on the right) create a ``net'' and successfully capture the blue agent. Thus, commanding one's swarm to surround an opposing agent, or to trap an opposing agent along the sides of the environment, are more likely to successfully capture the opposing agent than driving one's swarm to be directly on top of the opposing agent. The dark shaded areas of the environment shown in the middle figure represent areas the red swarm can capture blue agents in. Darker shaded regions are more likely to lead to captures than lighter shaded regions.}
	\label{fig:capture-agents}
	\vspace{-1em}
\end{figure*}

\subsection{The Agents}
Our virtual agents are modeled as second order, 2-D point masses.
Agents are spawned in one of four corners of the virtual ROS environment. 
Each agent is given a random initial altitude, distinct from all other agents in the environment, at which it remains for the entire game to avoid collisions.
Agents subscribe to ROS topics that players publish commands to. 
In our demonstrated examples, each agent receives commands via touchscreen interface (discussed further in Section~\ref{sec:touchscreen-interface}), runs its own real-time, decentralized ergodic control algorithm (discussed further in Section~\ref{sec:derg}), and receives information from members of the team it is currently on. Each agent runs their controls at 10 Hz.

\subsection{The Virtual ROS Environment}\label{sec:ros}
Our virtual ROS environment is rendered in a \verb|[0,1]|$^2$ grid in RViz. The game visualization shows areas of the environment in which a player's swarm can capture agents on the opposing team. These areas are denoted by black shading on the background of the environment. A ROS game engine node enforcing the rules of the game (discussed in Section~\ref{sec:game-rules}) sends information to another ROS node that controls the game visualization. This information determines each agent's color (representing the team they are on) at each time step. The game visualization also renders a decaying trajectory history of each agent so players can see where members of the opposing swarm have recently been. The game visualization and game engine nodes both run at 30 Hz.

\subsection{Rules for Capturing Agents}\label{sec:game-rules}
The time-averaged trajectories of each swarm dictates areas of the environment in which opposing agents can be captured. The pop-out in
Figure \ref{fig:capture-agents} shows how agents are captured. In order to create vaguely Go-like structures and motivate the concept of surrounding an opposing swarm, the time-averaged trajectories of the swarm over the whole environment (a \verb|50x50| grid) are averaged over smaller neighborhoods represented by \verb|5x5| sub-grids centered at each grid cell in the environment. Only the outer rows and columns of these sub-grids are used to calculate the ``value'' of each neighborhood. 

Thus, regions of the environment that are completely surrounded by a swarm, or that are surrounded by a swarm and one or more of the four boundaries of the environment, have higher potential to capture opposing agents than areas of the environment a swarm is directly positioned over. At each time step in the game, if an opposing agent enters into a neighborhood containing a capture value greater than 75\% of the maximum value over all neighborhoods, it will be captured. 
When an agent is captured, it immediately becomes controlled by the opposing player. The captured agent's command is updated to the last command published by the player controlling the new team it is now on. The captured agent also begins contributing to the areas its new team can capture opposing agents in through its trajectory history. If the new team the agent is on is running a decentralized control algorithm, the captured agent will also begin communicating with its teammates via ROS topics.  

Players have a clear visual representation of where their swarm can capture opposing agents via the game visualization in RViz. Areas where their swarms are likely to capture opposing agents are shaded in black, while areas with a low or no chance of capture are shaded in white. The visual representation helps the players build intuition about the game rules. The density of one's own swarm has no bearing on its safety; a player must ensure the safety of their swarm through commands driving their swarm away from areas the opposing swarm is surrounding. 

\section{Our Default System for Real-Time Human-Swarm Control}\label{sec:system-overview}

This section describes our default human-swarm control system, which consists of a touchscreen interface and decentralized ergodic control framework we have developed for demonstrating our game benchmark. Our system is scale and permutation-invariant, which enables players to command their swarm in real-time and respond to changes in the environment by re-specifying behavior for their swarm. 

\subsection{Touchscreen Interface}
\label{sec:touchscreen-interface}
\begin{figure}[t]
    \vspace{1em}
	\centering
	\includegraphics[width=3.25in]{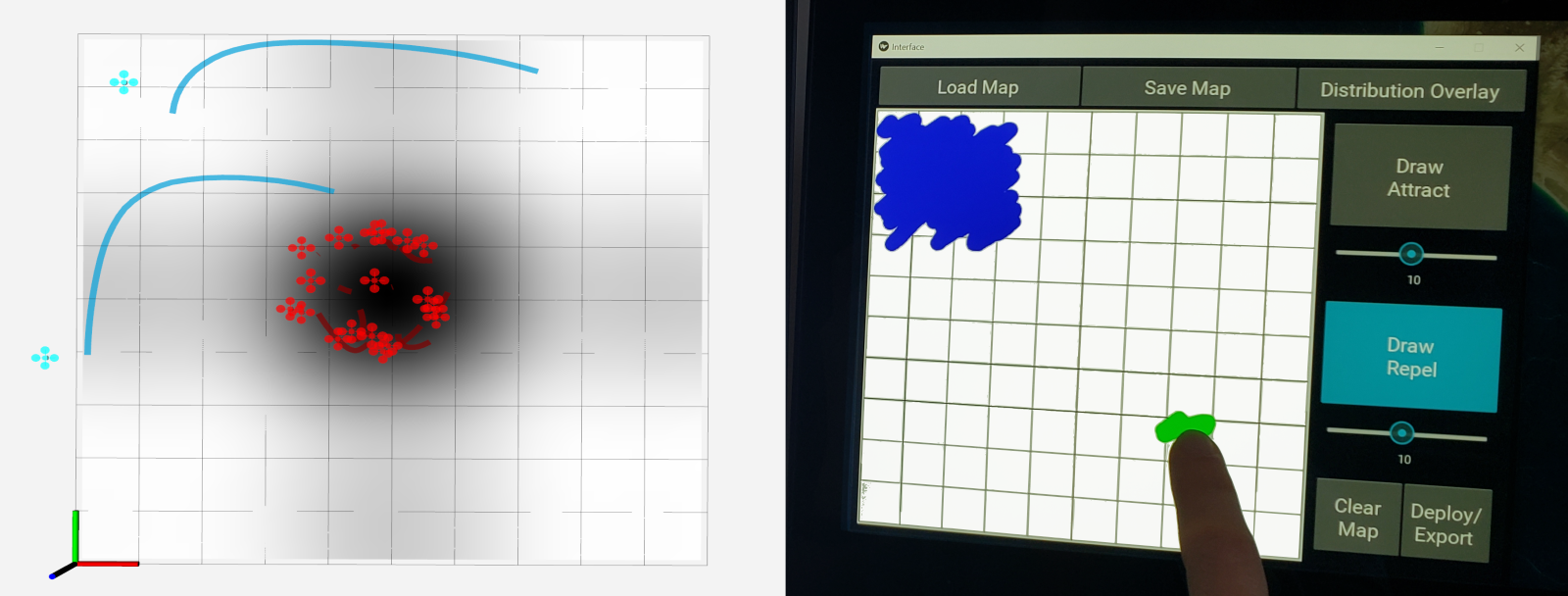}
	\caption{\textbf{Touchscreen Interface:} The virtual ROS environment (left) next to our touchscreen interface with a target distribution drawn (right). Players can draw areas of high interest (which will attract the agents) in blue and areas of low interest (which will repel the agents) in green. When a player is satisfied with the target distribution they have drawn, they can send the target distribution to their team. Players can clear the map to draw a new distribution. Players can also modify an existing target distribution by drawing new targets on top of old targets. The touchscreen's drawing area directly maps to the virtual ROS environment's area. The darker shaded regions in the virtual ROS environment figure on the left correspond to areas in which the red swarm can capture members of the blue swarm (darker shaded regions are more likely to lead to captures than lighter shaded regions).}
	\label{fig:touchscreen-interface}
    \vspace{-1em}
\end{figure}

Figure \ref{fig:touchscreen-interface} shows our system's touchscreen interface for sending player commands to the swarm. We created our touchscreen interface using Kivy---an open source Python framework for developing graphical applications with touch capabilities. Our touchscreen interface works on any PC or tablet running Ubuntu 18.04 LTS or Windows 10 with Tkinter, Python3, and OpenCV. Players can control their swarm by specifying distributions with hand-gestures on a touchscreen tablet or with a mouse on PC; these distributions specify what their team of agents should do by interpreting the path of the gesture as a spatial distribution. 

Players draw their distributions on a 2-D, top-down rendering of the virtual ROS environment. Players can select one of two input distribution types for areas of the environment: ``Attract'' (denoted in blue) which agents will converge to and spend more time in, and ``Repel'' (denoted in green) which agents will avoid. New target distributions can be continually overlaid on top of previous ones, which enables quick updates to previous distributions. Players' target distributions are smoothed out with a Gaussian filter. They are then scaled in value according to the resolution and size of the virtual ROS environment before being sent via ROS websocket to the swarm agents.

\subsection{Decentralized Ergodic Control}\label{sec:derg}
Player target distribution specifications provided via the touchscreen are transformed into low-level swarm commands through decentralized ergodic control. Decentralized ergodic control (described in detail in previous works \cite{abraham_decentralized_2018}, \cite{mavrommati_real-time_2017}, \cite{mathew_metrics_2011}) provides a natural framework for specifying density-based swarm objectives, and enables a player to quickly and flexibly specify behavior for their swarm in response to the opposing swarm's behavior. 

To define decentralized ergodic control, we start by introducing the dynamics of our system. Consider a set of $N$ agents with state $x(t) = \left[ x_1(t)^\top, x_2(t)^\top, \ldots, x_N(t)^\top\right]^\top :
    \mathbb{R}^+ \to \mathbb{R}^{n N}$. From work in \cite{abraham_decentralized_2018}, we define the dynamics of the collective multi-agent system as
    \begin{align} \label{eq:collective-dynamics}
        \dot{x} & = f(x,u) = g(x) + h(x) u \nonumber\\
        & = \begin{bmatrix}
        g_1(x_1) \\
        g_2(x_2) \\
        \vdots \\
        g_N(x_N)
        \end{bmatrix} +
         \begin{bmatrix}
        h_1(x_1) & \ldots & 0\\
        \vdots& \ddots & \\
        0 & & h_N(x_N)
        \end{bmatrix} u,
    \end{align}
where $g(x)$ is the free, unactuated dynamics of the multi-agent system and $h(x)$ is the system's dynamic control response. We seek to find a set of controls $u$ for the multi-agent system that 
minimizes the system's ergodic metric $\mathcal{E}$ with respect to some player target specification $\phi (s)$ where $s \in \mathbb{R}^2$ is a point in the game environment.

The ergodic metric $\mathcal{E}$ provides a way to calculate the ``difference'' between player target specifications  $\phi (s)$ and past agent trajectories \cite{mathew_metrics_2011} (which are represented as $c_k$ values using Fourier decompositions)---similar to the Kullback-Leibler divergence for comparing two distributions. 

We can use both Fourier decompositions to calculate the ergodic metric $\mathcal{E}$ (introduced in \cite{mathew_metrics_2011}):
\begin{equation}
\mathcal{E}(x(t)) = q \sum\limits_{k \in \mathbb{N}^\nu} \Lambda_k (c_k - \phi_k)^2 
\end{equation}
where $q \in \mathbb{R^+}$ is a scalar weight, $\Lambda_k=(1+||k||^2)^{\frac{\nu+1}{2}}$ is a weight on the frequency coefficients, $c_k$ is the Fourier decomposition of the agents' trajectories in a player's swarm over a fixed time horizon, and $\phi_k$ is the Fourier decomposition of the player's target specification $\phi (s)$ \cite{mathew_metrics_2011}.  

We can determine the ergodic metric's sensitivity to different control inputs $u$ by differentiating the ergodic metric with respect to a control application duration $\lambda$ at some optimal application time $\tau$. This results in the costate equation 
\begin{equation}
 \dot{\rho} = -\mathcal{E}(x(t)) \frac{\partial F_k}{\partial x} - \frac{\partial \Phi}{\partial x} - \frac{\partial f}{\partial x} \rho (t) 
\end{equation}
where $\Phi$ is the exponential control barrier function that keeps the agents in the operating environment, $F_k$ is the cosine basis function, and $f(x,u)$ is the system dynamics.

We can then write an unconstrained optimization problem
\begin{equation}
J = \int^{t_i+T}_{t_i}\frac{\partial \mathcal{E}}{\partial \lambda} \bigg|_\tau + \frac{1}{2} ||u_\star(t) - u_\text{def}(t)||^2_R \, dt
\end{equation}
where $R$ is a positive definite matrix that weights the control, $u_\star$ is the optimal control and $u_\text{def}$ is some default control. The default control $u_\text{def}$ could be that the agent moves forward at its current velocity. In this paper, $u_\text{def}$ is zero. The control that minimizes this objective $J$ is:
\[u_\star(t) = -R^{-1} \frac{\partial f}{\partial u}^T \rho(t) + u_\text{def}(t)\]
which is calculated and applied at every time step to the player's team of agents for the player's current target specification input. 
\begin{figure}[tb]
    \vspace{1em}
	\centering
	\includegraphics[width=3.25in]{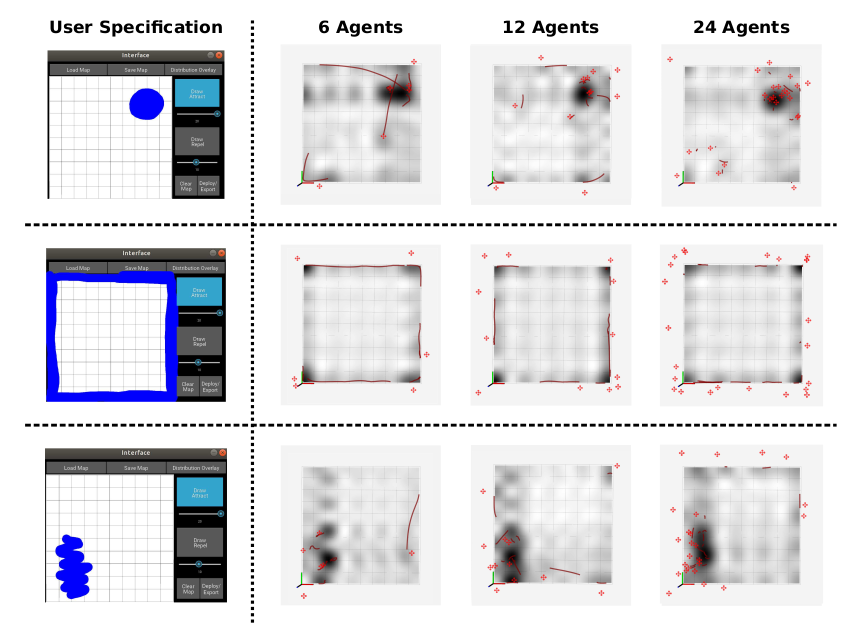}
	\caption{\textbf{Specification is Independent of Swarm Size with our Default Human-Swarm Control System:} In this example, the black and white shaded areas of the environment represent the time-averaged trajectories of the swarm (not capture areas based on the game rules), with black areas corresponding to areas the player's swarm's trajectories have spent more time in relative to white areas. The player's specifications are scale-invariant, since all three swarm sizes converge to the player target specifications. The player target specifications are also permutation-invariant, since the swarm will converge to the player's target specification from different permutations of the same set of agent positions.}
	\label{fig:scale-invariant-specifications}
\end{figure}

\subsection{Scale and Permutation-Invariance}
\label{sec:scale-invariant-permutation-invariant}
Our touchscreen interface and decentralized ergodic control algorithm enable our system to be scale and permutation-invariant with respect to player target distributions. Figure \ref{fig:scale-invariant-specifications} shows swarms containing 6, 12, and 24 agents responding to three different player target distributions specified through the touchscreen interface. The combination of our touchscreen interface and decentralized ergodic control algorithm produces scale-invariant swarm behavior, since all three swarm sizes converge to the three different target distributions. Our system also produces permutation-invariant swarm behavior, since the swarm will converge to the player's target distribution from different permutations of agent positions (i.e., the player's swarm contains 10 agents that are in different positions when the player specifies the target. If Agent 1, Agent 2, etc. swapped positions, the swarm would still converge to the player's target).

Our system's scale and permutation-invariance enables the player to plan swarm-level behavior instead of individual agent trajectories. For high-cadence scenarios like our game benchmark, as the number of agents the player controls increases and the time horizon the player has to plan decreases, the player cannot think strategically in terms of individual agents and their trajectories. Our system enables players to make strategic decisions based on the general positions and densities of their own swarm and their opponent's swarm. 

Furthermore, the decentralized nature of our ergodic control algorithm enables players to maintain their strategy (the most recent target distribution they sent to their swarm) regardless of how many agents are currently under their control. Even if the number of agents under their control is fluctuating, their swarm will still converge to their last target. These traits are advantageous for players planning tactics in high-cadence scenarios like our game benchmark. We elaborate on these traits in Section \ref{sec:swarm-tactics}.

\section{Swarm Tactics}\label{sec:swarm-tactics}
In this section, we describe an example ensemble tactic a player can deploy with our default human-swarm control system to beat an opponent in the game benchmark. We then discuss tactical considerations players may need to make during the game. We also discuss the shortcomings of other methods for human-swarm control and the challenges players may face when using these other methods to plan tactics in high-cadence scenarios like our game benchmark.

\subsection{Example Game Tactics}
\begin{figure}
    \vspace{1em}
	\centering
	\includegraphics[height=1.7in]{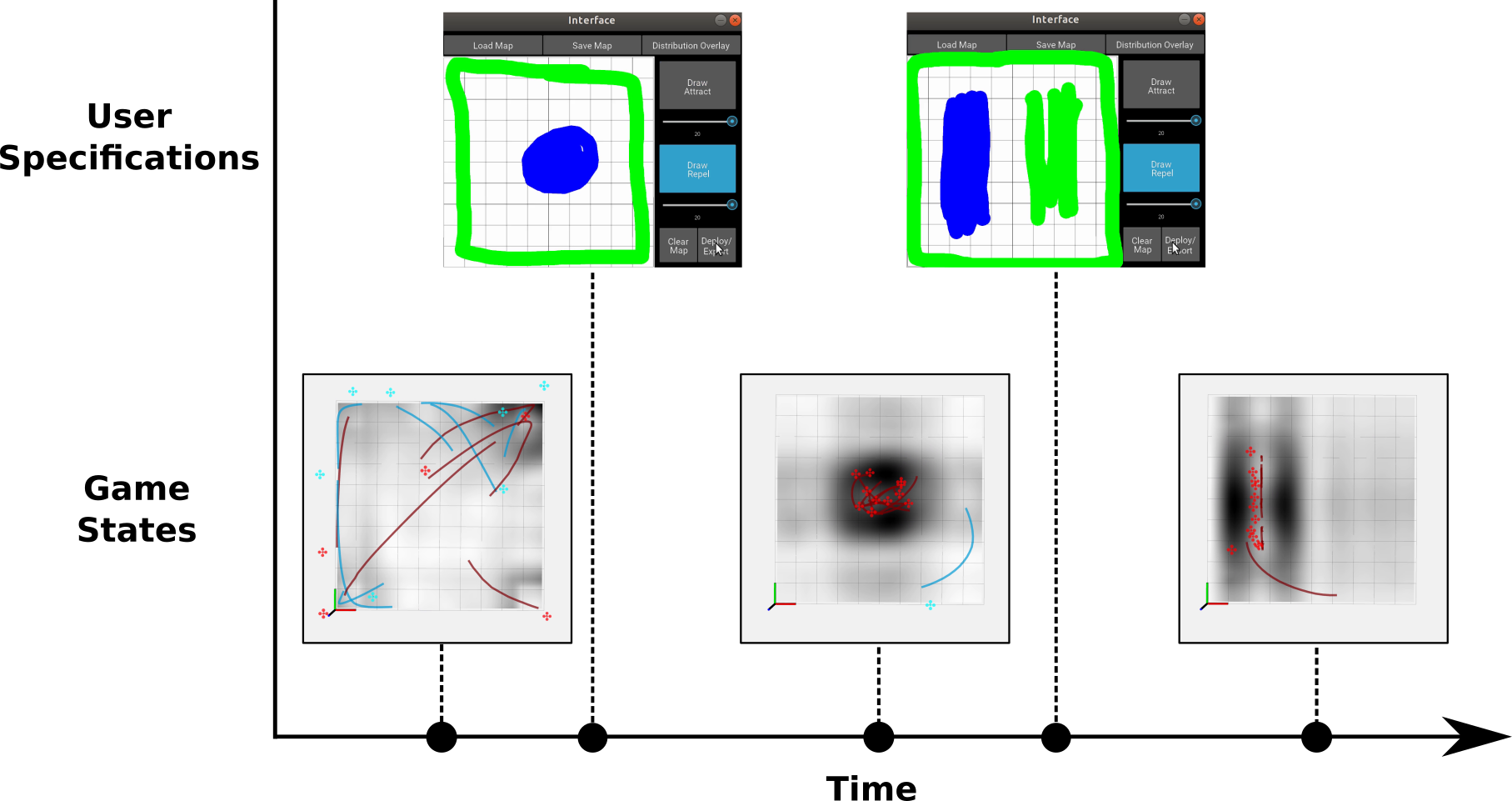}
	\caption{\textbf{Ensemble Game Tactics:} In the game state on the far left, it is difficult to discern any particular structure. Once the red player specifies a unimodal target distribution, the red agents converge to the center of the environment and capture opposing blue agents in the process. The red player then specifies their red agents to track the remaining blue agent along the left side of the environment to capture it and win the game. The dark shaded areas in the game state figures correspond to areas in which the red team can capture blue agents. The darker shaded regions are more likely to lead to captures than the lighter shaded regions.}
	\label{fig:swarm-tactics}
	\vspace{-1em}
\end{figure}

Figure \ref{fig:swarm-tactics} shows an example sequence of target specifications a player can use with our default human-swarm control system to capture agents on the opposing team. The nature of our game enables a player to quickly capture all the opposing agents (and win the game) with a well-timed sequence of commands. In the initial game state shown in Figure \ref{fig:swarm-tactics} on the far left, it is difficult to discern any particular structure in the agent positions. Once the red player specifies a unimodal distribution in the center of the environment with our interface, the red agents converge to a visible structure via decentralized ergodic control and capture opposing blue agents in the process. The red player then specifies their swarm to track the remaining blue agent as it circles along the left side of the environment. The red swarm then captures the blue agent and wins the game. 

This is one of many possible winning specification sequences players can make with our human-swarm control system (see multimedia attachment). Players can reason about how they can use the distribution of their swarm in the environment to create the most opportunities to capture opposing agents. Players can also reason about the current position of the opposing swarm and attempt to predict where the opposing swarm will be in subsequent time steps. 

\subsection{Player Tactic Considerations}
Since the rules of the game allow any agent to be captured regardless of how many of its team members are surrounding it, players might be wary of sending commands to their swarm that cause all of their agents to be in positions close to one another. While agents on the same team that are close together create areas of the environment that are more likely to capture agents on the opposing team (due to overlapping capture areas from the game rules), the opponent may maneuver their swarm in a way that enables them to simultaneously capture all the agents on the player's team and win the game. Thus, there are consequences to every maneuver a player makes; a player may maneuver to capture a small number of agents on the opposing team, only to find that their agents have now been steered into an area where they can be captured. 

\begin{figure}
    \vspace{1em}
	\centering
	\includegraphics[height=1.7in]{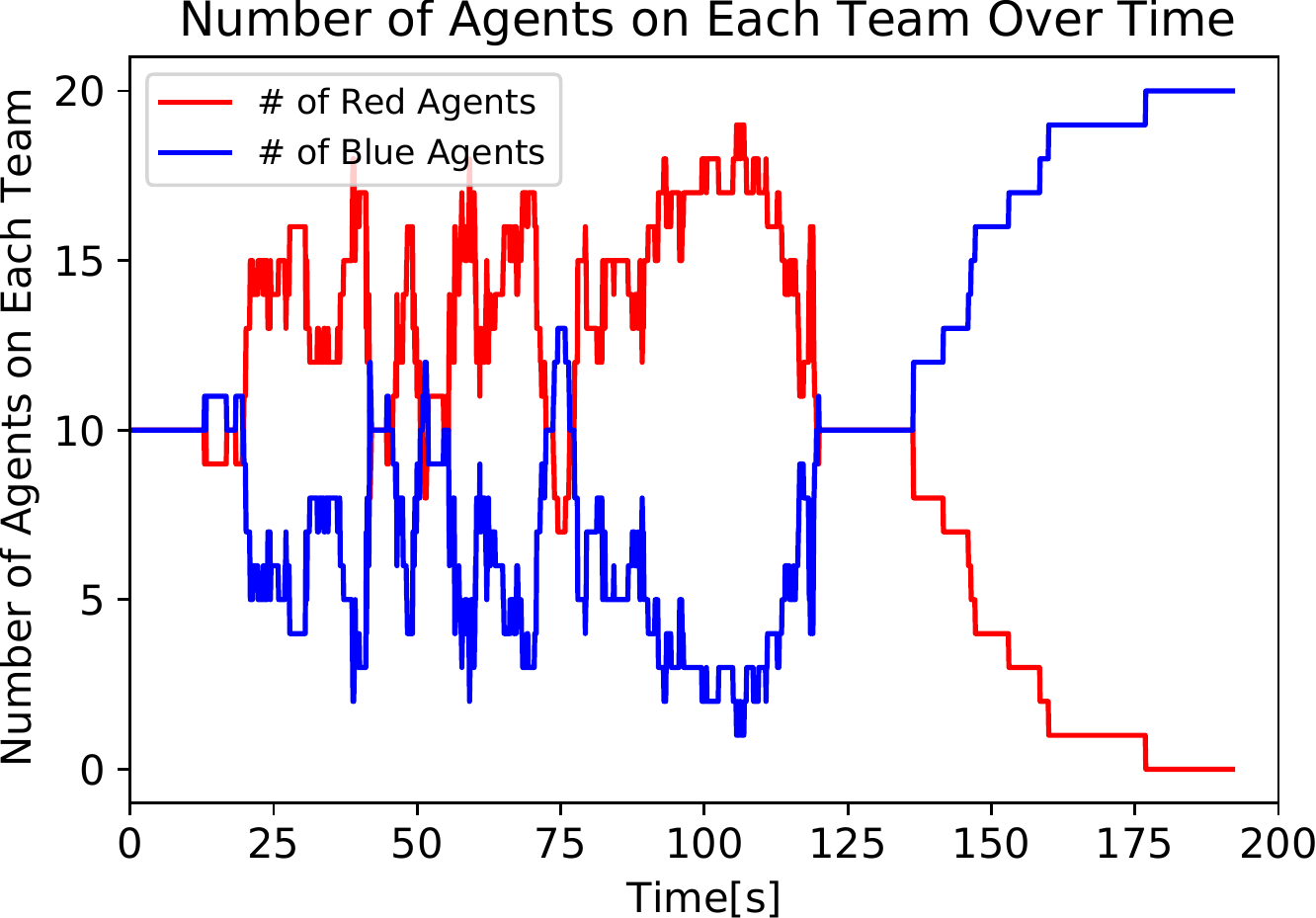}
	\caption{\textbf{Swarm Size Changes Rapidly During Game Play:} This figure shows the number of agents on each team over the course of an example game between two players. The game starts at approximately the 15s mark and ends right before the 200s mark. The number of agents on each team changes rapidly over the course of the game. Near the end of the game, the frequency of agent captures stabilizes as the blue player accumulates more total agents than the red player. The blue player eventually captures all the red agents and wins the game.}
	\label{fig:rapid-team-size-change}
\end{figure}

The dynamic, high-cadence nature of the game also requires the player to consider the amount of time they take to send commands to their swarm. Intricate commands (such as drawing detailed target specifications with our default human-swarm control system) may require the player to solely focus on creating their command (i.e., focusing on their touchscreen interface while they draw) instead of looking at the rapidly changing swarm positions in the environment. Figure \ref{fig:rapid-team-size-change} shows how the number of agents on each team fluctuates over the course of an example game. As seen in the figure, team sizes change rapidly over short periods of time. A brief advantage in number of agents under a player's control can quickly vanish in a matter of seconds. This rapid cadence may also affect player strategy in that players with larger swarms may be able to study the environment for emerging patterns in an opponent's play and take longer to specify new commands for their swarm since they can afford to lose agents without losing the game. Players down to the last member of their swarm may instead have to repeatedly specify commands that cause their agent to quickly move around the environment to create opportunities to capture opposing agents and rebuild the size of their swarm.  

We note here that it \emph{is possible} to recover from a dramatic ratio in numbers of agents between teams. One agent, controlled strategically, can be sufficient to build up team size again. Players may find that they need to play many games before they become adept at reasoning about their swarm's behavior as a whole. A naive player  may find that even an opponent swarm specified by a uniform coverage command across the domain---with no changes over time---is a challenging adversary.  

Various strategies for trapping and capturing agents on the opposing team may emerge for players from repeated game play. Player tactics may mirror some of those presented in other work in pursuit-evasion, such as \cite{pierson_intercepting_2017}, or in swarm tactics work such as \cite{day_responding_2018}. In some instances, players may find it useful to make the environment more chaotic by having their swarm spread out in all directions, which could disorient the opponent. For instance, players using our default human-swarm control system could crash their swarm into one side of the environment through ``wavefront'' attacks via specifically-drawn attraction and repulsion regions, which may quickly capture many opposing agents at high risk to the player losing their own agents.

\subsection{Challenges of Applying Traditional Swarm Control Methods to Game Tactics}
\label{sec:comparisons-to-nonscalable-methods}
Leader-follower methods such as those presented in \cite{goodrich_leadership_2012}, \cite{walker_human_2014} may not be able to perform the tactics described above because pre-designated leaders (or leaders elected at each time step) could be captured. Each team would have to re-appoint a leader from their available agents, which may not be possible with certain leader-election algorithms or methods that involve multiple leaders \cite{walker_control_2014} at high-cadence. Likewise, influencing the swarm by having the player take control of an individual agent or sub-team (such as in \cite{swamy_scaled_2020}) may also be infeasible since the agent the player is controlling could be captured. It may also be difficult for the player to select an individual agent or sub-team in a high-cadence environment due to cognitive load.

Formation-based control methods, such as those presented in \cite{michael_planning_2009}, \cite{hauri_multi-robot_2014}, \cite{hauert_flocking}, may limit how a player can specify different locations for sub-groups of their swarm to converge to for strategic purposes. It is not clear how the same multimodal maneuvers performed by drawing density specifications with our default system can be achieved with swarm formation control methods that may require fixed formations to be determined beforehand. Some of these formation-based methods that are non-decentralized may also be affected by communications failure or agents changing teams.

Both leader-follower and formation-based control methods, however, do have advantages over our default system in the different types of automatic swarm-response behavior they can enable. For instance, leader-follower methods could enable a player to create automatic multi-modal behaviors for elected leaders in their swarm to perform in response to different opposing swarm configurations during game play. Formation-based methods could enable a player to create a library of swarm formations they found useful for strategic purposes that they could then deploy at any instant during the game. Such automatic behaviors would not be possible with our default system (in its current form). 

\begin{figure*}[htb]
    \vspace{1em}
	\centering
	\includegraphics[width=5.0in]{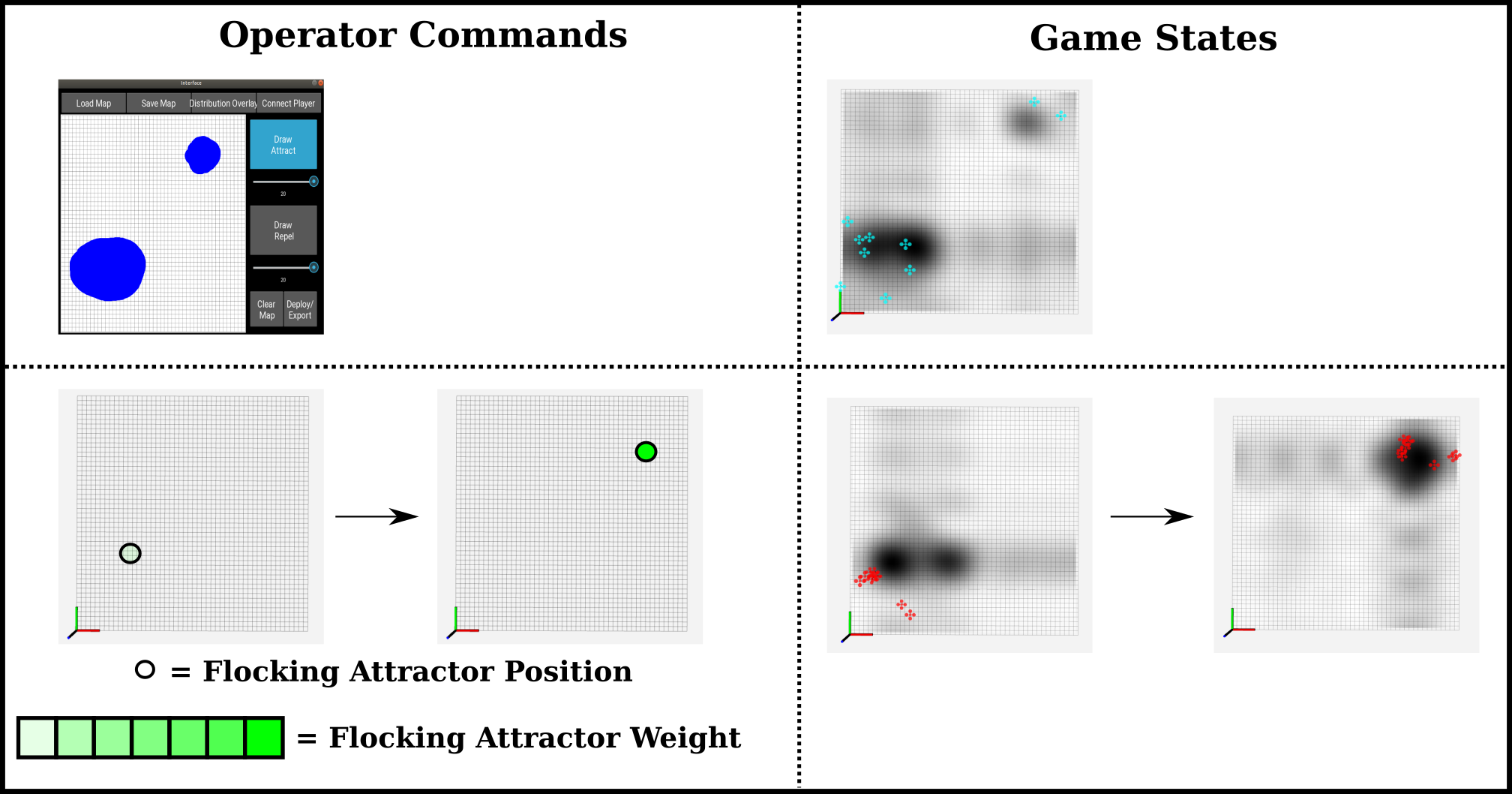}
	\caption{\textbf{Our Default Method for Human-Swarm Control vs. Flocking:} This figure compares how our system for human-swarm control and a flocking-based method converge to a bimodal target (with a more diffuse target in the bottom left corner and a more compact target in the upper right). The top row shows the results for our default system for human-swarm control. The player specifies the bimodal target all at once, drawing a larger circle on their interface for the more diffuse target and a smaller circle for the more compact target. The bottom row shows the results for the flocking-based method. The player has to specify each of the targets in a sequence instead of all at once. The player also has to specify both the flocking attractor positions (denoted by circles) and the attraction weights (denoted by green color saturation) for each of these targets to achieve the desired level of compactness.}
	\label{fig:flocking-vs-us}
\end{figure*}

Figure \ref{fig:flocking-vs-us} compares our default method for human-swarm control to a flocking-based control algorithm (adapted from work in \cite{hauert_flocking}). 
In the figure, the player specifies a bimodal target for their swarm to converge to. Our system enables players to specify both targets at once, while flocking requires both targets to be specified in sequence, one at a time (denoted by the circles in the figure). The desired ``compactness'' of the player's swarm at each target would also have to be specified through an attractor weight parameter (green color saturation). On the other hand, our system enables the player to draw a larger region for more diffuse coverage and a smaller region for more compact coverage in a single specification. 
Although our default system can perform some tactical maneuvers (such as the one above) in fewer specifications than the flocking-based method, the flocking-based method is still scale and permutation-invariant (like our default system) and may prove superior in other scenarios. For instance, specifying a single attractor position and weight to capture opposing agents in a particular area of the game environment may take a player less time than drawing an attractor region with our touchscreen interface.

Thus, while the system we have used to demonstrate the game benchmark contains many advantages over more traditional methods for swarm control, we only make those arguments in specific cases of specific algorithms; other researchers implementing their own systems for real-time human-swarm control in this game benchmark will lead to better comparisons and faster development of capable algorithms. The game benchmark is a useful opportunity for the swarm and human-robot interaction communities to determine what traits are necessary for these systems to have to succeed in high-cadence scenarios. We conclude with what some of these traits are in the next section. 

\section{Discussion}
\label{sec:discussion}
We have designed a game benchmark for assessing human-swarm control algorithms and interfaces in a real-time, high-cadence scenario. We demonstrated our game benchmark scenario using a default human-swarm control system that was scale and permutation-invariant. We provided discussion on example tactics players can employ, tactical considerations players may need to make while playing our game benchmark, and compared our default system to other methods for human-swarm control.

Based on our demonstrations, scale and permutation-invariance are important characteristics for algorithms and interfaces operating in high-cadence scenarios like our game benchmark. It is difficult to deploy tactics at high-cadence without algorithms and interfaces that scale to teams of different sizes and enable players to specify locations for their swarm to converge to without having to keep track of unique agent locations or which agents have been captured or newly acquired. Instead of reasoning about individual agents, players must be able to reason about their and their opponent's swarms as a whole.

A specific advantage of our default system for human-swarm control is that the touchscreen interface does not require player swarm specifications to be pre-determined. Different players can draw different distributions on the touchscreen that represent the same ``tactic'' at a higher level (i.e., trapping opposing agents by drawing some type of shape).
Players who are interested in using techniques from machine learning to learn possible tactics (that humans could deploy) for swarm systems operating in high-cadence scenarios may want to develop control algorithms and interfaces (with or without touch-based modalities) that afford machine learning agents the same specification flexibility as our system during the learning process. Scale and permutation-invariance may also be necessary traits for these algorithms and interfaces to make tactic learning feasible (especially if techniques from reinforcement learning are used). 

Future work could develop a virtual adversary to compete against human players in this benchmark scenario. We envision this adversary as an opponent benchmark that other researchers could use to test their real-time human-swarm control systems against. Then, the performance of real-time human-swarm control systems could be compared via number of wins and duration of game-play. Finally, the game benchmark could be extended to conduct human subject testing to empirically determine what types of strategies human players employ. Biometric data could be collected from players as they play the game to determine how much cognitive load they are experiencing and if certain changes to the human-swarm control algorithm or interface produce more or less cognitive load.




\section*{Acknowledgments}
This material is based on work supported by the United States National Science Foundation grant CNS 1837515 and by the 
Defense Advanced Research Projects Agency (DARPA) OFFSET SPRINT grant HR00112020035. The views, opinions, and/or findings expressed are those of the authors, and should not be interpreted as representing the views or policies of either agency. Author Joel Meyer was supported by a National Defense Science and Engineering Graduate Fellowship.



\bibliographystyle{IEEEtran}

\bibliography{2022CASEbibv6}

\begin{thebibliography}{10}
\providecommand{\url}[1]{#1}
\csname url@rmstyle\endcsname
\providecommand{\newblock}{\relax}
\providecommand{\bibinfo}[2]{#2}
\providecommand\BIBentrySTDinterwordspacing{\spaceskip=0pt\relax}
\providecommand\BIBentryALTinterwordstretchfactor{4}
\providecommand\BIBentryALTinterwordspacing{\spaceskip=\fontdimen2\font plus
\BIBentryALTinterwordstretchfactor\fontdimen3\font minus
  \fontdimen4\font\relax}
\providecommand\BIBforeignlanguage[2]{{%
\expandafter\ifx\csname l@#1\endcsname\relax
\typeout{** WARNING: IEEEtran.bst: No hyphenation pattern has been}%
\typeout{** loaded for the language `#1'. Using the pattern for}%
\typeout{** the default language instead.}%
\else
\language=\csname l@#1\endcsname
\fi
#2}}

\bibitem{durantin_using_2014}
G.~Durantin, J.-F. Gagnon, S.~Tremblay, and F.~Dehais,
  ``\BIBforeignlanguage{en}{Using near infrared spectroscopy and heart rate
  variability to detect mental overload},''
  \emph{\BIBforeignlanguage{en}{Behavioural Brain Research}}, vol. 259, pp.
  16--23, Feb. 2014.

\bibitem{olfati-saber_flocking_2006}
R.~Olfati-Saber, ``Flocking for multi-agent dynamic systems: algorithms and
  theory,'' \emph{IEEE Transactions on Automatic Control}, vol.~51, no.~3, pp.
  401--420, Mar. 2006.

\bibitem{howard_mobile_2002}
A.~Howard, M.~J. Matarić, and G.~S. Sukhatme, ``\BIBforeignlanguage{en}{Mobile
  {Sensor} {Network} {Deployment} using {Potential} {Fields}: {A}
  {Distributed}, {Scalable} {Solution} to the {Area} {Coverage} {Problem}},''
  in \emph{\BIBforeignlanguage{en}{Distributed {Autonomous} {Robotic} {Systems}
  5}}, 2002, pp. 299--308.

\bibitem{michael_planning_2009}
N.~Michael and V.~Kumar, ``\BIBforeignlanguage{en}{Planning and {Control} of
  {Ensembles} of {Robots} with {Non}-holonomic {Constraints}},''
  \emph{\BIBforeignlanguage{en}{The International Journal of Robotics
  Research}}, vol.~28, no.~8, pp. 962--975, Aug. 2009.

\bibitem{chen_distributed_2021}
J.~Chen, R.~Sun, and H.~Kress-Gazit, ``Distributed {Control} of {Robotic}
  {Swarms} from {Reactive} {High}-{Level} {Specifications},'' in \emph{{IEEE}
  17th {International} {Conference} on {Automation} {Science} and {Engineering}
  ({CASE})}, Aug. 2021.

\bibitem{szwaykowska_collective_2015}
K.~Szwaykowska, L.~M.-y.-T. Romero, and I.~B. Schwartz, ``Collective {Motions}
  of {Heterogeneous} {Swarms},'' \emph{IEEE Transactions on Automation Science
  and Engineering}, vol.~12, no.~3, pp. 810--818, July 2015.

\bibitem{zhao_self-adaptive_2018}
H.~Zhao, H.~Liu, Y.-W. Leung, and X.~Chu, ``Self-{Adaptive} {Collective}
  {Motion} of {Swarm} {Robots},'' \emph{IEEE Transactions on Automation Science
  and Engineering}, vol.~15, no.~4, pp. 1533--1545, Oct. 2018.

\bibitem{sampedro_flexible_2016}
C.~Sampedro, H.~Bavle, J.~L. Sanchez-Lopez, R.~A.~S. Fernández,
  A.~Rodríguez-Ramos, M.~Molina, and P.~Campoy, ``A flexible and dynamic
  mission planning architecture for {UAV} swarm coordination,'' in \emph{2016
  {International} {Conference} on {Unmanned} {Aircraft} {Systems} ({ICUAS})},
  June 2016.

\bibitem{diaz-mercado_distributed_2015}
Y.~Diaz-Mercado, S.~G. Lee, and M.~Egerstedt, ``Distributed dynamic density
  coverage for human-swarm interactions,'' in \emph{2015 {American} {Control}
  {Conference} ({ACC})}, July 2015.

\bibitem{prabhakar_ergodic_2020}
A.~Prabhakar, I.~Abraham, A.~Taylor, M.~Schlafly, K.~Popovic, G.~Diniz,
  B.~Teich, B.~Simidchieva, S.~Clark, and T.~Murphey,
  ``\BIBforeignlanguage{en}{Ergodic {Specifications} for {Flexible} {Swarm}
  {Control}: {From} {User} {Commands} to {Persistent} {Adaptation}},''
  \emph{\BIBforeignlanguage{en}{Robotics: Science and Systems}}, June 2020.

\bibitem{tsykunov_swarmtouch_2019}
E.~Tsykunov, R.~Agishev, R.~Ibrahimov, L.~Labazanova, A.~Tleugazy, and
  D.~Tsetserukou, ``{SwarmTouch}: {Guiding} a {Swarm} of {Micro}-{Quadrotors}
  {With} {Impedance} {Control} {Using} a {Wearable} {Tactile} {Interface},''
  \emph{IEEE Transactions on Haptics}, vol.~12, no.~3, pp. 363--374, July 2019.

\bibitem{karavas_hybrid_2017}
G.~K. Karavas, D.~T. Larsson, and P.~Artemiadis, ``A hybrid {BMI} for control
  of robotic swarms: {Preliminary} results,'' in \emph{{IEEE}/{RSJ}
  {International} {Conference} on {Intelligent} {Robots} and {Systems}
  ({IROS})}, Sept. 2017.

\bibitem{pinciroli_buzz_2016}
C.~Pinciroli and G.~Beltrame, ``Buzz: {An} extensible programming language for
  heterogeneous swarm robotics,'' in \emph{{IEEE}/{RSJ} {International}
  {Conference} on {Intelligent} {Robots} and {Systems} ({IROS})}, Oct. 2016.

\bibitem{lee_haptic_2011}
D.~Lee, A.~Franchi, P.~R. Giordano, H.~I. Son, and H.~H. Bülthoff, ``Haptic
  teleoperation of multiple unmanned aerial vehicles over the internet,'' in
  \emph{{IEEE} {International} {Conference} on {Robotics} and {Automation}},
  May 2011.

\bibitem{preiss_crazyswarm_2017}
J.~A. Preiss, W.~Honig, G.~S. Sukhatme, and N.~Ayanian,
  ``\BIBforeignlanguage{en}{Crazyswarm: {A} large nano-quadcopter swarm},'' in
  \emph{\BIBforeignlanguage{en}{{IEEE} {International} {Conference} on
  {Robotics} and {Automation} ({ICRA})}}, May 2017.

\bibitem{michael_grasp_2010}
N.~Michael, D.~Mellinger, Q.~Lindsey, and V.~Kumar, ``The {GRASP} {Multiple}
  {Micro}-{UAV} {Testbed},'' \emph{IEEE Robotics Automation Magazine}, vol.~17,
  no.~3, pp. 56--65, Sept. 2010.

\bibitem{pickem_robotarium_2017}
D.~Pickem, P.~Glotfelter, L.~Wang, M.~Mote, A.~Ames, E.~Feron, and
  M.~Egerstedt, ``The {Robotarium}: {A} remotely accessible swarm robotics
  research testbed,'' in \emph{{IEEE} {International} {Conference} on
  {Robotics} and {Automation} ({ICRA})}, May 2017.

\bibitem{kate_simbeeotic_2012}
B.~Kate, J.~Waterman, K.~Dantu, and M.~Welsh, ``Simbeeotic: {A} simulator and
  testbed for micro-aerial vehicle swarm experiments,'' in \emph{{ACM}/{IEEE}
  11th {International} {Conference} on {Information} {Processing} in {Sensor}
  {Networks} ({IPSN})}, Apr. 2012.

\bibitem{chung_live-fly_2016}
T.~H. Chung, M.~R. Clement, M.~A. Day, K.~D. Jones, D.~Davis, and M.~Jones,
  ``\BIBforeignlanguage{en}{Live-fly, large-scale field experimentation for
  large numbers of fixed-wing {UAVs}},'' in
  \emph{\BIBforeignlanguage{en}{{IEEE} {International} {Conference} on
  {Robotics} and {Automation} ({ICRA})}}, May 2016.

\bibitem{wang_game-theoretic_2021}
M.~Wang, Z.~Wang, J.~Talbot, J.~C. Gerdes, and M.~Schwager, ``Game-{Theoretic}
  {Planning} for {Self}-{Driving} {Cars} in {Multivehicle} {Competitive}
  {Scenarios},'' \emph{IEEE Transactions on Robotics}, vol.~37, no.~4, Aug.
  2021.

\bibitem{shah_multi-agent_2019}
K.~Shah and M.~Schwager, ``\BIBforeignlanguage{en}{Multi-agent {Cooperative}
  {Pursuit}-{Evasion} {Strategies} {Under} {Uncertainty}},'' in
  \emph{\BIBforeignlanguage{en}{Distributed {Autonomous} {Robotic} {Systems}}},
  2019.

\bibitem{shah_grape_2019}
------, ``\BIBforeignlanguage{en}{{GRAPE}: {Geometric} {Risk}-{Aware}
  {Pursuit}-{Evasion}},'' \emph{\BIBforeignlanguage{en}{Robotics and Autonomous
  Systems}}, vol. 121, Nov. 2019.

\bibitem{pierson_intercepting_2017}
A.~Pierson, Z.~Wang, and M.~Schwager, ``\BIBforeignlanguage{en}{Intercepting
  {Rogue} {Robots}: {An} {Algorithm} for {Capturing} {Multiple} {Evaders}
  {With} {Multiple} {Pursuers}},'' \emph{\BIBforeignlanguage{en}{IEEE Robotics
  and Automation Letters}}, vol.~2, no.~2, pp. 530--537, Apr. 2017.

\bibitem{pierson_controlling_2018}
A.~Pierson and M.~Schwager, ``\BIBforeignlanguage{en}{Controlling
  {Noncooperative} {Herds} with {Robotic} {Herders}},''
  \emph{\BIBforeignlanguage{en}{IEEE Transactions on Robotics}}, vol.~34,
  no.~2, pp. 517--525, Apr. 2018.

\bibitem{day_responding_2018}
M.~Day, L.~Strickland, E.~Squires, K.~DeMarco, and C.~Pippin,
  ``\BIBforeignlanguage{en}{Responding to unmanned aerial swarm saturation
  attacks with autonomous counter-swarms},'' in
  \emph{\BIBforeignlanguage{en}{Ground/{Air} {Multisensor} {Interoperability},
  {Integration}, and {Networking} for {Persistent} {ISR} {IX}}}, May 2018.

\bibitem{abraham_decentralized_2018}
I.~Abraham and T.~D. Murphey, ``Decentralized {Ergodic} {Control}:
  {Distribution}-{Driven} {Sensing} and {Exploration} for {Multiagent}
  {Systems},'' \emph{IEEE Robotics and Automation Letters}, vol.~3, no.~4, Oct.
  2018.

\bibitem{mavrommati_real-time_2017}
A.~Mavrommati, E.~Tzorakoleftherakis, I.~Abraham, and T.~D. Murphey,
  ``\BIBforeignlanguage{en}{Real-{Time} {Area} {Coverage} and {Target}
  {Localization} using {Receding}-{Horizon} {Ergodic} {Exploration}},''
  \emph{\BIBforeignlanguage{en}{IEEE Transactions on Robotics}}, vol.~34,
  no.~1, pp. 62--80, Aug. 2017.

\bibitem{mathew_metrics_2011}
G.~Mathew and I.~Mezić, ``\BIBforeignlanguage{en}{Metrics for ergodicity and
  design of ergodic dynamics for multi-agent systems},''
  \emph{\BIBforeignlanguage{en}{Physica D: Nonlinear Phenomena}}, vol. 240, no.
  4-5, pp. 432--442, Feb. 2011.

\bibitem{goodrich_leadership_2012}
M.~A. Goodrich, S.~Kerman, and S.-Y. Jung, ``\BIBforeignlanguage{en}{On
  {Leadership} and {Inﬂuence} in {Human}-{Swarm} {Interaction}},''
  \emph{\BIBforeignlanguage{en}{Association for the Advancement of Artificial
  Intelligence (AAAI)}}, p.~6, 2012.

\bibitem{walker_human_2014}
P.~Walker, S.~Amirpour~Amraii, N.~Chakraborty, M.~Lewis, and K.~Sycara, ``Human
  control of robot swarms with dynamic leaders,'' in \emph{{IEEE}/{RSJ}
  {International} {Conference} on {Intelligent} {Robots} and {Systems}}, Sept.
  2014.

\bibitem{walker_control_2014}
P.~Walker, S.~Amirpour~Amraii, M.~Lewis, N.~Chakraborty, and K.~Sycara,
  ``Control of swarms with multiple leader agents,'' in \emph{{IEEE}
  {International} {Conference} on {Systems}, {Man}, and {Cybernetics} ({SMC})},
  Oct. 2014.

\bibitem{swamy_scaled_2020}
G.~Swamy, S.~Reddy, S.~Levine, and A.~D. Dragan, ``Scaled {Autonomy}:
  {Enabling} {Human} {Operators} to {Control} {Robot} {Fleets},'' in
  \emph{{IEEE} {International} {Conference} on {Robotics} and {Automation}
  ({ICRA})}, May 2020.

\bibitem{hauri_multi-robot_2014}
S.~Hauri, J.~Alonso-Mora, A.~Breitenmoser, R.~Siegwart, and P.~Beardsley,
  ``\BIBforeignlanguage{en}{Multi-{Robot} {Formation} {Control} via a
  {Real}-{Time} {Drawing} {Interface}},'' in
  \emph{\BIBforeignlanguage{en}{Field and {Service} {Robotics}}}, 2014,
  vol.~92, pp. 175--189.

\bibitem{hauert_flocking}
S.~Haubert, S.~Leven, M.~Varga, F.~Ruini, A.~Cangelosi, J.-C. Zufferey, and
  D.~Floreano, ``Reynolds {Flocking} in {Reality} with {Fixed-Wing} {Robots}:
  {Communication} {Range} vs. {Maximum} {Turning} {Rate},'' in \emph{IEEE/RSJ
  International Conference on Robots and Systems}, Sept. 2011, pp. 5015--5020.

\end{thebibliography}

\end{document}